\documentclass[sigconf]{acmart}
\AtBeginDocument{%
  }

\setcopyright{acmlicensed}
\copyrightyear{2018}
\acmYear{2018}
\acmDOI{XXXXXXX.XXXXXXX}

\acmConference[Conference acronym 'XX]{Make sure to enter the correct
  conference title from your rights confirmation emai}{June 03--05,
  2018}{Woodstock, NY}
\acmISBN{978-1-4503-XXXX-X/18/06}

\begin{document}

\renewcommand{\labelenumii}{\arabic{enumi}.\arabic{enumii}}
\renewcommand{\labelenumiii}{\arabic{enumi}.\arabic{enumii}.\arabic{enumiii}}
\renewcommand{\labelenumiv}{\arabic{enumi}.\arabic{enumii}.\arabic{enumiii}.\arabic{enumiv}}

\title{MM-PhyRLHF: Reinforcement Learning
Framework for Multimodal Physics
Question-Answering}


\author{Janak Kapuriya$^*$}
\email{kapuriya22032@iiitd.ac.in}
\affiliation{%
  \institution{IIIT Delhi}
  \state{Delhi}
  \country{India}
}

\author{Chhavi Kirtani$^*$}
\email{chhavi18229@iiitd.ac.in}
\affiliation{%
  \institution{IIIT Delhi}
  \state{Delhi}
  \country{India}
}

\author{Apoorv Singh}
\email{apoorv17027@iiitd.ac.in}
\affiliation{%
 \institution{IIIT Delhi}
  \state{Delhi}
  \country{India}
}

\author{Jay Saraf}
\email{jay20438@iiitd.ac.in}
\affiliation{%
  \institution{IIIT Delhi}
  \state{Delhi}
  \country{India}
}

\author{Naman Lal}
\email{namanlal.lal92@gmail.com}
\affiliation{%
  \institution{IIIT Delhi}
  \state{Delhi}
  \country{India}
}

\author{Jatin Kumar}
\email{jatin20206@iiitd.ac.in}
\affiliation{%
  \institution{IIIT Delhi}
  \state{Delhi}
  \country{India}
}

\author{Adarsh Raj Shivam}
\email{adarsh20274@iiitd.ac.in}
\affiliation{%
  \institution{IIIT Delhi}
  \state{Delhi}
  \country{India}
}

\author{Astha Verma}
\email{asthav@iiitd.ac.in}
\affiliation{%
  \institution{IIIT Delhi}
  \state{Delhi}
  \country{India}
}

\author{Avinash Anand}
\email{avinasha@iiitd.ac.in}
\affiliation{%
  \institution{IIIT Delhi}
  \state{Delhi}
  \country{India}
}

\author{Rajiv Ratn Shah}
\email{rajivratn@iiitd.ac.in}
\affiliation{%
  \institution{IIIT Delhi}
  \state{Delhi}
  \country{India}
}

% \author{Roger Zimmermann}
% \email{rogerz@comp.nus.edu.sg}
% \affiliation{%
%   \institution{National University of Singapore}
%   \city{Singapore}
%   \country{Singapore}
% }

\renewcommand{\shortauthors}{Kapuriya et al.}

\begin{abstract}
% This research introduces MM-PhyQA, an innovative dataset for Indian high school physics education, addressing the gap in domain-specific resources. The study emphasizes the significance of Large Multimodal Models (LMMs) in understanding complex educational content through textual and visual data. It further explores the integration of the Reinforcement Learning from Human Feedback (RLHF) methodology to enhance the human-like problem-solving abilities of LMMs. The RLHF approach, incorporating human feedback into the learning process improves the model's problem-solving skills, truthfulness, and reasoning capabilities, minimizing the hallucinations in the answers and improving the quality as opposed to using vanilla-supervised fine-tuned models. We employ the LLaVA open-source model to answer multimodal physics MCQs and compare the performance with and without using RLHF.
% \\
% \newline
% (Alternative) Added why we experimented with RLHF and The multimodal property of PhyQA: \newline

Recent advancements in LLMs have shown their significant potential in tasks like text summarization and generation. Yet, they often encounter difficulty while solving complex physics problems that require arithmetic calculation and a good understanding of concepts. Moreover, many physics problems include images that contain important details required to understand the problem's context. We propose an LMM-based chatbot to answer multimodal physics MCQs. For domain adaptation, we utilize the MM-PhyQA dataset comprising Indian high school-level multimodal physics problems. To improve the LMM's performance, we experiment with two techniques, RLHF (Reinforcement Learning from Human Feedback) and Image Captioning. In image captioning, we add a detailed explanation of the diagram in each image, minimizing hallucinations and image processing errors. We further explore the integration of Reinforcement Learning from Human Feedback (RLHF) methodology inspired by the ranking approach in RRHF to enhance the human-like problem-solving abilities of the models. The RLHF approach incorporates human feedback into the learning process of LLMs, improving the model's problem-solving skills, truthfulness, and reasoning capabilities, minimizing the hallucinations in the answers, and improving the quality instead of using vanilla-supervised fine-tuned models. We employ the LLaVA open-source model to answer multimodal physics MCQs and compare the performance with and without using RLHF.
\def\thefootnote{*}\footnotetext{These authors contributed equally to this work}
% To tackle this, we introduce MM-PhyQA dataset comprising of Indian high school level multimodal physics problems. This dataset will help to address the gap in domain-specific understanding of the LLMs. To further improve the performance of LLMs, we explore the integration of Reinforcement Learning from Human Feedback (RLHF) methodology inspired by the ranking approach in RRHF to enhance the human-like problem-solving abilities of the models. The RLHF approach incorporates human feedback into the learning process of LLMs, improving the model's problem-solving skills, truthfulness, and reasoning capabilities, minimizing the hallucinations in the answers, and improving the quality as opposed to using vanilla-supervised fine-tuned models. We employ the LLaVA open-source model to answer multimodal physics MCQs (free style questions?) and compare the performance with and without using RLHF.
\end{abstract}

\begin{CCSXML}
<ccs2012>
   <concept>
       <concept_id>10010405.10010432.10010441</concept_id>
       <concept_desc>Applied computing~Physics</concept_desc>
       <concept_significance>500</concept_significance>
       </concept>
 </ccs2012>
\end{CCSXML}

\ccsdesc[500]{Applied computing~Physics}

\keywords{Large Language Models, Large Multimodal Models, Reinforcement Learning, Physics, Educational AI}


\maketitle

\section{Introduction}
 % <current LLMs capabilities. Includes different types of data inputs primarily textual and visual. importance of multimodal. why multimodal is important in question-answering systems. why inclusion of images is important in physics questions> \\
% The current generation of Large Multimodal models showcases substantial advancements in the areas of multimodal data processing, contextual reasoning, and pattern recognition. The approach represents a significant leap and intertwines different types of data inputs which are primarily textual and visual content to develop a more effective and extensive model. The importance of multimodal lies in the ability to interpret and understand various data types. This incorporation is very crucial in the field of question-answering systems, where the inclusion of images can immediately increase the understandability of the question exponentially and result in more accuracy. In the case of Physics questions, including images is crucial as it provides significant information not present in the textual data.
% \\
% Alternative of the above paragraph:\\

% In-general trend of LLMs currently
The current generation of LLMs like Chat-GPT and GPT4 have shown impressive performance in tasks like text generation, translation, summarization, sentiment analysis, and question answering, enabling more accurate and contextually relevant language understanding and generation. With newer research, these models have also shown impressive performance in domain-specific tasks through fine-tuning.
Moreover, newer capabilities of handling multimodal input data are making strides in the academic landscape, exemplified by GPT-4V, Gemini, and open-source LLaVA models that can process and generate content across multiple modalities, including images, audio, and video.\\
% \cite{sun2023aligning}
% Need for AI in the education domain and how LLM helps
% About LLM applications in question-answering in the education domain

Combining the LLM's capability of data processing, text generation, contextual reasoning, and pattern recognition with its capabilities of handling multiple modalities has made it a popular option for educational question-answering. LLMs can serve as 24/7 virtual assistants for answering student queries in various subjects, thus enhancing understanding and clearing doubts outside the traditional classroom setting. These LLMs can revolutionize exam preparation by creating well-rounded question-answering systems with domain-specific assistance and instant clarifications. LLMs can also tailor content according to each student's learning style, needs, and preferences by analyzing the way a student interacts and performs. \\

Recent research has portrayed LLM's ability to retrieve and provide accurate information from vast knowledge bases, essential for educational question-answering systems \cite{li2023adapting, xiao2023evaluating} noted positive feedback received from middle school teachers about using LLMs in education stating that LLMs significantly reduce the time and cost of obtaining diverse and personalized learning materials. \cite{abbasiantaeb2024let} underscore LLMs' capability of simulating high-quality student-teacher interaction while covering a broader aspect of the given topic, suggesting LLMs' potential to enhance the depth and breadth of educational content. Hence, developing open-source educational question-answering models using LLMs is promising. While some work has been done in the field of multimodal question-answering in science, maths, etc., this work is fresh in the Indian context, especially for high-quality physics problems at the level of Indian competitive examinations like JEE and NEET.
\\

% Multiple research [1](Li et al, 2024) [2](Xiao, et al, 2023) has been conducted to evaluate LLMs performance in the education domain, discussing its current potential and future improvements. [2](Xiao, et al, 2023) 

% These LLMs can revolutionize exam preparation by creating well-rounded question-answering systems with domain-specific assistance and instant clarifications.

% The current generation of Large Multimodal models showcases substantial advancements in the areas of multimodal data processing, contextual reasoning, and pattern recognition, showing impressive capabilities in handling multiple modalities.
\cite{li2023adapting} underline the importance of developing LLMs capable of handling multimodal information for educational problems that include text and images, which is crucial for creating more comprehensive and effective educational tools. Moreover, \cite{xiao2023evaluating} highlighted the importance of feedback collection as a method to enhance domain-adaptation of subjects and offer further improvement of LLMs in the educational domain. As noted by \cite{li2023adapting}, it is imperative to enhance the multimodal capabilities of currently available LLMs to create a more well-rounded question-answering system. Multi-modal processing problems such as hallucinations where the model perceives information that is non-existent and non-sensical, inability to integrate modalities to understand the context and the significance of all modalities about each other, and domain adaptation issues are still persistent. Improving the multimodal capabilities of models is crucial, especially in the field of question-answering, where the inclusion of images can exponentially increase the understandability of the question and result in more accuracy. Furthermore, as noted by \cite{anand2023context, anand2023gec, xiao2023evaluating} LLMs' contextual reasoning and capability of generating more human-like responses in the educational domain can be improved. \\

MICoT-QA \cite{anand2024mm} compared the performance of fine-tuning open-source LLMs, while additionally exploring the technique of Chain-of-Thought Prompting. To tackle the problems faced by LLMs in multimodal processing as well as domain-specific reasoning as discussed above, we propose improvements to MICoT-QA by utilizing two techniques, Reinforcement Learning and Human Feedback (RLHF) and Image Captioning. \\

For every image in the training dataset, we add a detailed explanation of the diagram to provide the model with extra context helping it avoid hallucinations. We utilize the Infi-MM captioning model to generate captions for every image, before passing these images along with the accompanying question and answers to the fine-tuning pipeline. \\

To enhance the model's reasoning capabilities and align their response to a more humanized version is where Reinforcement Learning from Human Feedback(RLHF) plays an important role.
% Inspired by the ranking process in RRHF, we train our model on different variants of responses for the same sample question ranked on the basis of human-generated feedback.
At the core of RLHF lies preference data which involves rating and comparing various responses generated by different models to the same prompt. The model is further trained on this preference data and thereafter multiple reinforcement learning algorithms like Proximal Policy Optimization (PPO) \cite{schulman2017proximal} and Direct Preference Optimization (DPO) \cite{rafailov2024direct} are applied to fine-tune the LMMs to optimize the rewards. RLHF technique has been proven to boost LLMs reasoning by bridging the gap between human intuition and AI and generating more rational and human-aligned solutions (ref: Boosting LLMs Physics Reasoning with Human-AI Guided Reinforcement Learning Conclusion section). \\
\\
% we generate 6 different answers for the same sample question using 6 models namely, llava\_13b, llava\_1\_5\_7b, llava\_1\_5\_13b, llava\_1\_5\_13b\_lora\_large, and GPT4, and \?. These responses are then ranked by the level of preference using Gemini Pro model. Once ranked, we pair the responses in the following
% We explore the Chain of Thoughts and Reinforcement Learning and Human Feedback methodologies to further improve the performance of the model. We showcase how the reasoning capability of LLMs improves significantly by providing chain of thought prompting, where the model is provided with a few chain of thought demonstrations as exemplars during prompting.
% To enhance the multimodal models further and align their response to a more humanized response is where Reinforcement Learning from Human Feedback(RLHF) plays an important role. RLHF integrates reinforcement learning techniques with human guidance. It involves the training of the model using human-generated feedback to guide the learning process in order to generate a humanized response. 
The three stages of RLHF are:
\begin{enumerate}
    \item Collecting human feedback to rank multiple responses to the same questions based on the quality of their reasoning.
    \item Training the reward model: The Reward Model (RM) is trained methodologically using our dataset MM-PhyQA.
    \item Refining the policy: An iterative process where the policy keeps getting updated based on new feedback and changes.
\end{enumerate}
\noindent
% \vspace{1mm}
% collecting human feedback
% *add here*
% training the reward model 
% 	The Reward Model (RM) module is trained methodologically using our dataset Phyllm
% refining the policy
% 	An iterative process where the policy keeps getting updated based on the new feedback 
% 	and changes.
Overall, our contributions to this paper are as follows:
\begin{enumerate}
  \item Enhance the MM-PhyQA dataset by using the image captioning methodology. This benefits the LLM by reducing hallucinations and adding more context to the problem.
  \item Applying the RLHF technique to encourage more human-like responses to complex problems.
  \item   Compare results of the LLM-based physics question answering between the following experiment settings:
  \begin{enumerate}
    \item Fine Tune LLaVa Models using (Sample Question and Answer, Image, Caption) with RLHF technique
    \item Fine Tune LLaVa Models using (Sample Question and Answer, Caption) with RLHF technique
    \item Fine Tune LLaVa Models using (Sample Question and Answer, Image) with RLHF technique
\end{enumerate}
  
\end{enumerate}
% 1. Enhance the MM-PhyQA dataset by using the image captioning methodology. This benefits the LLM by reducing hallucinations and adding more context to the problem. \\
% 2. Applying RLHF technique to encourage more human-like responses to complex problems. \\
% 3. Compare results of the LLM-based physics question answering between the following experiment settings: \\

% RLHFs focus on the humanized responses and COTs ability to mimic human cognitive processes while problem-solving enables our model to not only provide accurate solutions but also in a format that is easily understandable by students. Developing upon the improvements made using RLHF, the Chain of Thoughts (COT) method emerges as a major factor for Large Multimodal Model enhancement. COT substantially improves the model's problem-solving capability which still is a standing problem with Large Multimodal models. It enables the model to prompt a series of interconnected thoughts, copying the human cognitive process. This will improve the model's ability to tackle complex physics problems making the model more efficient and effective in problem solving through step-by-step reasoning. The incorporation of COT not only makes the model more precise and accurate but also provides a very transparent reasoning path making it easier for the students to follow and understand the working of the solution. Furthermore, the sequential process of COT helps in error detection and correction, increasing the model's learning and adaptation capabilities.
\section{Related Work}

% the datasets available for physics
% Large Language Models (LLMs) have made substantial progress in complex problem-solving within education, as evidenced by the use of multimodal datasets like GeoQA, TQA, ChartQA, MMQA, and ScienceQA. GeoQA (Chen et al., 2021) and TQA (Welbl et al., 2017) demonstrate how these models can handle middle school curricula with a mix of images, text, and tables. ChartQA (Masry et al., 2022) and MMQA (Gupta et al., 2018) further illustrate this versatility. However, despite these achievements, as Lu et al. (2022) in ScienceQA suggest, there's an ongoing need for LLMs that can effectively address high-school-level complexity, pushing AI's educational capabilities further.

% growing capability of LLMs to handle multiple modalities (related multimodal llms present)
\subsection{Vision LLMs}
Recent advancements have led to the creation of Vision Language Models, models capable of processing and interpreting visual information along with textual information. Flamingo has been designed to leverage visual and language inputs as prompts, showing remarkable few-shot performance in visual question-answering tasks. Models like GPT4 \cite{achiam2023gpt}, LLaVA series \cite{liu2024visual}, and MiniGPT4 \cite{zhu2023minigpt} introduced the method of visual instruction tuning to enhance the instruction-following abilities of VLLMs, meaning training the model to interpret and process visual cues in conjunction with textual instructions. VisionLLM \cite{wang2024visionllm}, Kosmos-2 \cite{peng2023kosmos}, Qwen-VL \cite{bai2023qwen} have been improved by using visual grounding capabilities, enabling them to perform tasks such as describing specific regions within an image (region description) and identifying the location of objects or elements within an image (localization), making them more adept at tasks that require nuanced understanding of the visual elements. PaLM-E \cite{driess2023palm} and EmbodiedGPT \cite{mu2024embodiedgpt} models are designed to interact with and navigate through physical or simulated environments, understanding visual inputs in the context of actions and interactions, which represents significant strides in adapting VLLMs for embodied applications.

% The integration of multimodal modalities into Large Language Models (LLMs) like GPT4-V, Flamingo (Alayrac et al., 2022), and BLIP-2 (Li et al.) has notably enhanced their effectiveness in question-answering scenarios. These models blend textual and visual data, fostering a deeper understanding of complex queries. Similarly, advancements in PaLM-E (Driess et al., 2023) and LLaVA (Liu et al., 2023), along with foundational models like CLIP (Radford et al., 2021) and Vicuna (Peng et al., 2023), demonstrate the growing versatility and capability of LLMs in providing contextually rich and accurate responses.

% high school physics education work

% RLHF (Related works of the QA in the context of RLHF)
\subsection{RLHF}
In addressing the limitations of traditional LLM approaches, the emergence of Reinforcement Learning from Human Feedback (RLHF) has been pivotal, as noted in research by \cite{xie2023olagpt}. This method, initially focused on tasks such as text summarization \cite{stiennon2020learning} and question answering \cite{nakano2021webgpt}, has evolved to improve general-purpose language models. The RLHF domain has expanded further with Reinforcement Learning from AI Feedback (RLAIF) by \cite{lee2023rlaif}, addressing the scalability challenges and the need for quality human labels, as analyzed by \cite{zheng2024judging}. Additionally, UltraFeedback \cite{cui2023ultrafeedback} represents a significant advancement in ranking diverse human and AI preferences, highlighting the ongoing evolution and adaptation in language model development. In the context of question-answering systems, \cite{kirk2023understanding, goel2023advancements} presents an in-depth analysis of the impact of RLHF on LLMs by evaluating each stage of an RLHF pipeline separately, showcasing the importance of RLHF in improving Out-Of-Distribution performance suggesting that RLHF models are more adaptable to a wider variety of scenarios not seen during training. \cite{nakano2021webgpt} explored the benefit of RLHF in browser-assisted question answering with human feedback on the performance in question answering. We utilize the PPO method for the RLHF implementation in our application. The Proximal Policy Optimization (PPO) Algorithm was introduced by \cite{schulman2017proximal} by ChatGPT creator OpenAI's scientist as a method to implement policy updates and reward models in RLHF. It proposes a mechanism that restricts the policy update steps, preventing them from being too large and potentially destabilizing the training process, and unlike standard policy gradient methods that perform one update per data sample, PPO allows for several epochs of minibatch updates on the same set of sampled data, offering a balance between performance and ease of implementation.
 % on how to obtain these files.)

% Related works of Captioning in the context of QA
\subsection{Image Captioning}
To address the limitations of multimodal processing of LLMs and to reduce the hallucinations of the models, image captioning has been effective as noted by research conducted by \cite{salaberria2023image} and \cite{zhang2024mathverse}. \cite{salaberria2023image} conducted experiments comparing the performance of available LMMs with multimodal text and image input to the performance of LLMs with text-only input accompanied by image captioning instead of the image in Visual Question Answering tasks. The research reported improved accuracy for LLMs given text-only image-captioned input, representing the benefit of providing the model with more context by adding image captions. \cite{zhang2024mathverse} explored image captioning for mathematical question answering by giving 6 varied versions of a problem, each offering varying degrees of multi-modality information as image captions. They report that most MLLMs struggle to understand mathematical diagrams and heavily rely on textual data, signifying the importance of image captioning.

\subsection{Application in Education}
\cite{filippo2024future} studies the many applications of LLMs, including education. LLMs present opportunities for personalized learning materials, improved productivity, and increased accessibility. \cite{mitra2024elevating} developed an LLM-based student-facing assistant in the discussion forum and tested the performance in upper-division data science courses, suggesting the practicality of deploying LLMs as tutoring and question-answering assistants. \cite{prather2023robots} present an extensive literature review of the usage of LLMs in education, specifically in the computing domain, as an AI assistant for teaching coding \cite{puryear2022github}, answering questions on the course discussion boards \cite{jaipersaud2023decomposed}, providing feedback for programming assignments \cite{balse2023investigating, pankiewicz2023large}, and more. The advancements in Large Language Models (LLMs) within high-school physics education are highlighted in the research "Revolutionizing High School Physics Education: A Novel Dataset" \cite{anand2023revolutionizing}. This study showcases the creation of a specialized dataset from NCERT exemplar solutions, expanding from 766 to 7,983 physics questions. The primarily text-based dataset, formatted in JSON, demonstrates LLMs' potential in handling complex educational content, as evidenced by significant METEOR and BERTScore F1 metrics. This development is pivotal in LLM applications for high-school education, enriching the scope of multimodal learning environments. \cite{sapkota2024assessing} reported the limitations of ChatGPT in mathematical education tasks like question generation, suggesting a need for further improvement in domain adaptation and contextual understanding of LLMs.

% Factually augmented RLHF (Captioning + RLHF) - didn't find any related work for this

\section{Dataset}

We possess the MMPhy-QA dataset, which is an extensive compilation comprising a diverse range of multimodal multiple-choice physics questions along with their corresponding answers. This dataset has been meticulously curated to serve as a valuable resource for promoting a comprehensive understanding and assessment of physics principles. Within this dataset, the training section consists of 3700 samples, carefully selected to cover a broad spectrum of physics topics, while the test section comprises 676 samples, intentionally designed to evaluate the performance and adaptability of models.

In our efforts to further strengthen the robustness and effectiveness of our model, we strategically extracted 2000 instances from the training set to create a specialized preference dataset, tailored specifically for Reinforcement Learning with Human Feedback (RLHF) training. This strategic augmentation aims to leverage human feedback to refine and enhance the model's responses, thereby fostering deeper understanding and more refined reasoning abilities.
% finely-tuned model, which has been honed to capture the subtle nuances of the MMPhy-QA dataset. Subsequent to the inference process, we utilized Gemini Pro, 
Each of the 2000 selected samples underwent meticulous inference using models llava\_13b, llava\_1\_5\_7b, llava\_1\_5\_13b, llava\_1\_5\_13b\_lora\_large, and GPT4 to generate 5 responses. These 5 responses were then passed on to Gemini Pro, an advanced evaluation tool renowned for its sophisticated ranking algorithms and analytical capabilities. Leveraging the capabilities of Gemini Pro, we meticulously evaluated and ranked the generated responses, scrutinizing their overall coherence in reasoning and adherence to the provided prompts.

Based on the rankings obtained from Gemini Pro, we arranged the responses into paired sets, pairing the highest-ranked response with every other response. This meticulous pairing approach resulted in four distinct preference pairs per sample, effectively quadrupling the dataset's size and infusing it with diverse viewpoints and nuanced variations. Consequently, our preference dataset now comprises a comprehensive array of 8000 samples, each meticulously crafted to foster nuanced comprehension and facilitate robust model training.

To provide a deeper understanding of the structure and composition of the dataset, we have included a detailed diagram delineating the intricate relationships and pairings within the preference dataset. This graphical representation serves as a valuable tool, offering insights into the dataset's organizational framework and aiding in the comprehension of its complexities and subtleties.

\begin{figure*}[t]
    \centering
    \Description{Preference dataset Pipeline}
    \includegraphics[width=0.8\textwidth]{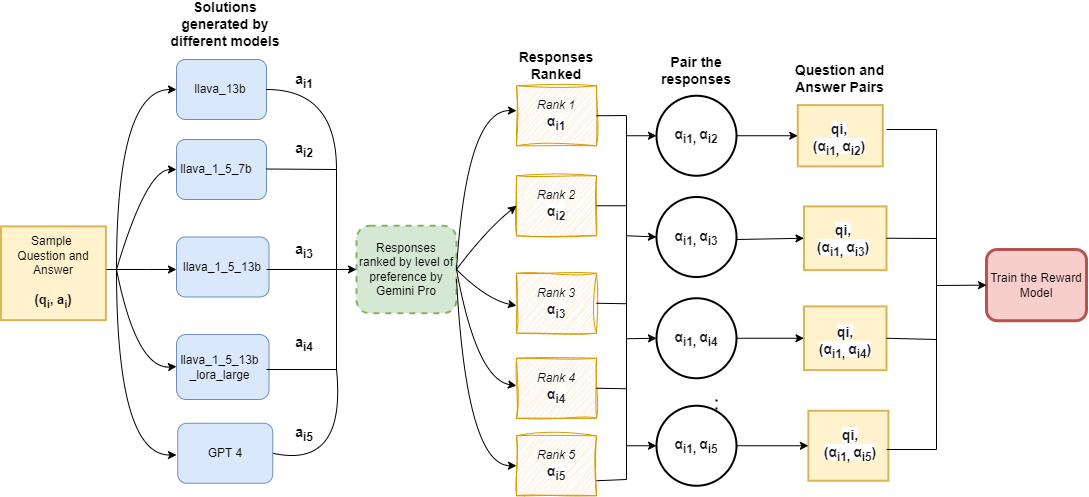}
    \caption{Preference dataset Pipeline}
    \label{fig:sample_cot_question}
\end{figure*}

% \begin{figure*}[t]
%     \centering
%     \includegraphics[width=0.8\textwidth]{RLHF_pipeline.pdf}
%     \caption{Preference dataset Pipeline}
%     \label{fig:sample_cot_question}
% \end{figure*}

\section{Proposed Methodology}

\subsection{Dataset}
We utilize the MM-PhyQA dataset from "MM-PhyQA: Multimodal Physics Question-Answering With Multi-Image CoT Prompting" \cite{anand2023revolutionizing, anand23mathify, anand2023sciphyrag, anand2023kg} for experimentation in this paper. MM-PhyQA addresses the lack of multimodal datasets that comprise physics questions and are catered to high school students. The dataset is geared toward competitive examinations conducted throughout India. 300 questions in the dataset were initially manually created, each sample consists of a question, four options, and a final answer along with an explanation. Further, ChatGPT was used for data augmentation by giving a prompt to create other variations of the text while ensuring that the meaning remained the same, bringing the total count of the questions in the dataset to 4500.

\subsection{Preference dataset creation for RLHF}

We employed the MM-PhyQA dataset to construct a preference dataset for the training of the Reward Model (RM) component. The dataset incorporates contributions from five open-source models: llava\_13b, llava\_1\_5\_7b, llava\_1\_5\_13b, llava\_1\_5\_13b\_lora\_large, and GPT4, each generating five distinct responses. These responses were subjected to an evaluation process using Gemini Pro, a sophisticated evaluation tool renowned for its advanced ranking algorithms and analytical prowess. Leveraging the capabilities of Gemini Pro, we conducted a thorough assessment and ranking of the responses, focusing on their coherence in reasoning and adherence to the provided prompts. The evaluation outcomes facilitated the organization of the responses into paired sets by pairing the highest-ranked response with each of the other responses. This method of pairing yielded four unique preference pairs per sample, effectively augmenting the dataset's volume and introducing a rich diversity of perspectives and subtle distinctions. As a result, the preference dataset was expanded to include a total of 8000 samples, each carefully curated to enhance nuanced understanding and support the effective training of the model.
% We use the MM-PhyQA dataset to further create preference dataset for training Reward Model (RM) module. This dataset
% draws upon the collective expertise of five open-source models, namely llava\_13b, llava\_1\_5\_7b, llava\_1\_5\_13b, llava\_1\_5\_13b\_lora\_large, and GPT4 to generate 5 responses. These 5 responses were then passed on to Gemini Pro, an advanced evaluation tool renowned for its sophisticated ranking algorithms and analytical capabilities. Leveraging the capabilities of Gemini Pro, we meticulously evaluated and ranked the generated responses, scrutinizing their overall coherence in reasoning and adherence to the provided prompts. Based on the rankings obtained from Gemini Pro, we arranged the responses into paired sets, pairing the highest-ranked response every other response. This meticulous pairing approach resulted in four distinct preference pairs per sample, effectively quadrupling the dataset's size and infusing it with diverse viewpoints and nuanced variations. Consequently, our preference dataset now comprises a comprehensive array of 6000 samples, each meticulously crafted to foster nuanced comprehension and facilitate robust model training.
\subsection{Fine-Tuning}
We use the MM-PhyQA dataset consisting of 4500 samples to fine-tune each of these pre-trained models: Llava\-1.5 7b, Llava\-1.5 13b, Llava\-1.5 13b lora large. The dataset is split into training and testing sets with a 70:30 ratio
% Following need to confirm
To optimize the training of these LLMs, we employed PEFT (Parameter Efficient Fine Tuning) \cite{liu2022few}. Additionally, the hyper-parameter configuration utilized for training the SFT module includes 4 epochs, a batch size of 8, a learning rate of 3e-4, and an Adam optimizer.
\subsection{RLHF Pipeline}
% We integrate RLHF technique in the training of our LLM to improve its contextual reasoning and align its responses with human preferences. We use the aforementioned Preference Dataset to train a Reward Model (RM). For this paper, we have chosen LLaVA2-13B as our Reward Model. 

We incorporate the Reinforcement Learning from Human Feedback (RLHF) methodology into the training process of our LLM to enhance its contextual reasoning capabilities and ensure its outputs are in line with human preferences. The Preference Dataset mentioned in section 4.2 serves as the foundation for training a Reward Model (RM), selected as LLaVA2-13B for this research. By training the RM on the preference dataset, it learns to predict the quality of a response based on the rankings. It utilizes the criteria of coherence in reasoning and adherence to initial prompts as established during the creation of the Preference Dataset to predict the anticipated reward for any response produced by the LLM.
% By using the criteria of coherence in reasoning and adherence to the provided prompts used while creating the preference dataset, this RM aims to predict the expected reward for any given response generated by the LLM.
After training the RM, we use it to apply the PPO algorithm to each of the fine-tuned LLMs. During this phase, the LLM generates responses that are evaluated by the RM, and it receives feedback in the form of predicted rewards from the RM. The process of generating responses, evaluating them with the RM, and updating the LLM using policy optimization is repeated iteratively.
% how many iterations?

\subsection{Image Captioning}
To enhance multimodal processing and reduce the hallucinations of the model, we add captions to every image in the training dataset during fine-tuning. We utilized Infi-MM \cite{liu2024infimm}, a novel architecture designed to recognize and understand detailed information in high-resolution images accurately, in our pipeline that takes input as the sample dataset MM-PhyQA and returns the samples accompanied with image captions. This modified dataset is then used for the fine-tuning of the pre-trained model, as described in section 4.3.

\section{Experiments}
Using the above methodology, we experiment to evaluate and compare the performance of physics question-answering LLM in the context of Indian high schools. We will integrate the RLHF and Image Captioning techniques with the novel method introduced by \cite{anand2023sciphyrag, goel2023advancements} which is multi-image chain-of-thought prompting to enhance LLM's performance for physics-based question-answering. By integrating RLHF and Image Captioning techniques, we aim to improve the accuracy and contextual reasoning of the LLMs and contribute to the advancement of educational chatbots.

We experiment with the following six training settings:

\begin{enumerate}
    \item Fine Tune LLaVa Models using (Sample Question and Answer, Image, Caption)
    \item Fine Tune LLaVa Models using (Sample Question and Answer, Caption)
    \item Fine Tune LLaVa Models using (Sample Question and Answer, Image)
    \item Fine Tune LLaVa Models using (Sample Question and Answer, Image, Caption) with RLHF technique integrated
    \item Fine Tune LLaVa Models using (Sample Question and Answer, Caption) with RLHF technique integrated
    \item Fine Tune LLaVa Models using (Sample Question and Answer, Image) with RLHF technique integrated
\end{enumerate}

By comparing the performance of the LLM across the above six different settings, we will be able to understand the extent of improvement Image Captioning and RLHF techniques provide.

% This research is an extension of our previous paper "MM-PhyQA: Multimodal Physics Question-Answering With Multi-
% Image CoT Prompting" (Anand, et al, 2024). In our previous work, we experimented with COT-Prompting where the model is provided with a few chain of thought demonstrations as exemplars during prompting. In this research, on top of the techniques used in our previous paper, we have added RLHF technique and image captioning. The performance evaluated through these new integrations will be compared with the results obtained previously.

Furthermore, we use the following three LLMs as our base model:
\begin{enumerate}
    \item Llava -1.5 7b
    \item Llava -1.5 13b
    \item Llava -1.5 13b LoRA large
\end{enumerate}

We experimented with different LoRA values in the case of the LLaVA 1.5 model. LoRA or Low-Rank Adaptation \cite{hu2021lora}, is a method to represent the weight changes during the training process in lower-ranked matrices. This is especially useful while fine-tuning general-purpose LLMs, as it speeds up the training process. A lower LoRA rank means fewer parameters are learned during the adaptation process, however, it results in a faster training process as well. We tested the 7b (7 billion) and 13b (13 billion) variants of LLaVA which correspond to the number of learning parameters. The different LLaVA configurations also formed the basis of our comparison of the performance of our technique.

\section{Results and Discussion}
Tables 1, 2, and 3 show the accuracy scores on the test dataset for the experiment settings 1-3 mentioned in section 5 - Experiments.
\begin{table}[H] 
\label{table:1}
\begin{tabular}{ |p{3cm}|p{2.5cm}|p{2.5cm}| }
 \hline
 \multicolumn{3}{|c|}{SFT Training Input: Text + Image} \\
 \hline
 Model & Accuracy & LoRA\\
 \hline
 Llava -1.5 7b & 53.3\% & 64  \\
  \hline
 Llava -1.5 13b & 52.7\% & 64 \\
  \hline
 Llava -1.5 13b lora large & 53.1\% & 128 \\
 \hline
\end{tabular}
\caption{Accuracy recorded when the training dataset contained only the Question and Answer text and the accompanying image. This experiment involves only Supervised Fine-Tuning.}
\end{table}

\begin{table}[H] 
\label{table:2}
\begin{tabular}{ |p{3cm}|p{2.5cm}|p{2.5cm}| }
 \hline
 \multicolumn{3}{|c|}{SFT Training Input: Text + Image + Caption} \\
 \hline
 Model & Accuracy & LoRA\\
 \hline
 Llava -1.5 7b & 82.52\% & 64 \\
  \hline
 Llava -1.5 13b & 83.28\% & 64 \\
  \hline
 Llava -1.5 13b lora large & 82.1\% & 128 \\
 \hline
\end{tabular}
\caption{Accuracy recorded when the training dataset contained only the Question and Answer text, the accompanying image, and the image's caption. This experiment involves only Supervised Fine-Tuning.}
\end{table}

\begin{table}[H] 
\label{table:3}
\begin{tabular}{ |p{3cm}|p{2.5cm}|p{2.5cm}| }
 \hline
 \multicolumn{3}{|c|}{SFT Training Input: Text + Caption} \\
 \hline
 Model & Accuracy & LoRA\\
 \hline
 Llava -1.5 7b & 66.95\% & 64 \\
  \hline
 Llava -1.5 13b & 64.0\% & 64 \\
  \hline
 Llava -1.5 13b lora large & 74.56\% & 128 \\
 \hline
\end{tabular}
\caption{Accuracy recorded when the training dataset contained only the Question and Answer text and the caption of the accompanying image (Image was not provided). This experiment involves only Supervised Fine-Tuning.}
\end{table}

\section{Conclusion}

This study investigates the impact of incorporating image captioning models on the multimodal reasoning capabilities of open-source multimodal large language models (LLMs). Our findings indicate that providing the LLM with only the question text and the image (Table 1) results in the poorest performance. In contrast, augmenting the input with an image caption generated by a captioning model, along with the question text and image (Table 2), significantly improves reasoning accuracy. Interestingly, we observe that excluding the image entirely and supplying only the caption and question text (Table 3) yields worse performance than the comprehensive setting in Table 2, where the model is provided with the question text, image, and caption.

Future work will extend this approach by leveraging Reinforcement Learning with Human Feedback (RLHF). By incorporating human preference knowledge, we aim to align the reasoning capabilities of multimodal LLMs more closely with human expectations, further enhancing their performance and applicability in real-world scenarios.

\bibliographystyle{ACM_Reference_Format}
\bibliography{software}

\end{document}